\documentclass[oneside,a4paper,onecolumn,11pt]{article}

\usepackage{fullpage}
\usepackage[left=2cm,top=2cm,bottom=2cm,right=2cm,includehead,nomarginpar,headheight=16pt]{geometry}
\usepackage[ruled, linesnumbered, vlined, commentsnumbered]{algorithm2e}
\usepackage{graphicx,subfig} 
\usepackage{amsfonts,amssymb,amsmath,amsthm,amsopn,mathtools}	
\usepackage{booktabs,diagbox,colortbl,multirow,tabularx,threeparttable,hhline}
\usepackage[listings,skins,breakable]{tcolorbox}

\usepackage{fancyhdr,fancyheadings,nopageno,lastpage} 
\usepackage{enumerate}
\usepackage[shortlabels]{enumitem}
\usepackage{csquotes}
\usepackage{authblk}
\usepackage{footnote}
\usepackage[colorlinks=true,citecolor=reference,linkcolor=brightmaroon,backref=page]{hyperref}
\usepackage{prettyref}
\usepackage{cite}
\usepackage{setspace}
\usepackage{color}
\usepackage{titlesec}
\usepackage{soul}
\usepackage{url}
\usepackage{bm}
\usepackage{xcolor}  
\usepackage{csquotes}
\usepackage{academicons}
\usepackage{fontawesome5}
\usepackage{pifont,ifsym,marvosym,manfnt} 

\usepackage{tikz}
\usepackage{pgfplots}
\usetikzlibrary{snakes}
\usepackage{rotating}

\def\hlinew#1{%
  \noalign{\ifnum0=`}\fi\hrule \@height #1 \futurelet
   \reserved@a\@xhline}
\makeatother

\usetikzlibrary{positioning,shapes,shadows,arrows,calc}
\tikzstyle{component}=[rectangle, draw=black, rounded corners, fill=blue!40, drop shadow, text centered, anchor=north, text=white, minimum height=1cm]
\tikzstyle{arrow}=[->, thick]

\pgfplotsset{compat=1.12}
\usetikzlibrary{intersections}
\usetikzlibrary{pgfplots.statistics}
\usepgfplotslibrary{fillbetween}

\fancypagestyle{plain}{%
    \fancyfoot[C]{}
    \fancyfoot[R]{\faLeanpub\ \, \thepage\ / \pageref{LastPage}}
    \fancyfoot[L]{\faBraille\ \textsc{COLALab Report} \faSlackHash\ \footnotesize\textifsym{2024006}}
}

\hypersetup{
    colorlinks=true,
    linkcolor=ultramarine,
    filecolor=magenta,
    urlcolor=ultramarine,
    pdftitle={Overleaf Example},
    pdfpagemode=FullScreen,
}

\setlist[itemize]{itemsep=0pt}
\setlist[enumerate]{itemsep=0pt}

\definecolor{red(munsell)}{rgb}{0.95, 0.0, 0.24}
\definecolor{navyblue}{RGB}{0, 0, 128}
\definecolor{myblue}{RGB}{34,31,217}
\definecolor{Gray}{gray}{0.9}
\definecolor{usccardinal}{rgb}{0.6, 0.0, 0.0}
\definecolor{ultramarine}{RGB}{0,32,96}
\definecolor{amber}{rgb}{1.0, 0.49, 0.0}
\definecolor{reference}{RGB}{4, 20, 110}
\definecolor{linenum}{RGB}{21, 127, 127}
\definecolor{amaranth}{rgb}{0.9, 0.17, 0.31}
\definecolor{brightmaroon}{rgb}{0.76, 0.13, 0.28}
\definecolor{lightpurple}{RGB}{242, 239, 246}
\definecolor{mycyan}{gray}{.7}

\newtheorem{definition}{Definition}

\newsavebox\newcaptionbox\newdimen\newcaptionboxwid
\titlespacing*{\paragraph}{0pt}{0.75ex plus 0.75ex minus 0.2ex}{1.25ex plus 0.2ex}

\long\def\@makecaption#1#2{
 \vskip 10pt
        \baselineskip 11pt
        \setbox\@tempboxa\hbox{#1. #2}
        \ifdim \wd\@tempboxa >\hsize
        \sbox{\newcaptionbox}{\small\sl #1.~}
        \newcaptionboxwid=\wd\newcaptionbox
        \usebox\newcaptionbox {\footnotesize #2}
        \else
          \centerline{{\small\sl #1.} {\small #2}}
        \fi}

\newtcolorbox{quotebox}{
    colback=lightpurple,
    colframe=black!75,
    boxrule=0pt,
    top=5pt,
    bottom=5pt,
    left=8pt,
    right=8pt,
    arc=8pt,
    boxsep=0pt,
    toptitle=2pt,
    bottomtitle=2pt,
    fonttitle=\bfseries,
}

\SetCommentSty{mycommfont}

\newcommand{\pref}{\prettyref}

\newrefformat{fig}{Fig.~\ref{#1}}
\newrefformat{tab}{Table~\ref{#1}}
\newrefformat{sec}{Section~\ref{#1}}
\newrefformat{alg}{Algorithm~\ref{#1}}
\newrefformat{property}{Property~\ref{#1}}
\newrefformat{theorem}{Theorem~\ref{#1}}
\newrefformat{definition}{Definition~\ref{#1}}
\newrefformat{corollary}{Corollary~\ref{#1}}
\newrefformat{lemma}{Lemma~\ref{#1}}
\newrefformat{conj}{Conjecture~\ref{#1}}
\newrefformat{def}{Definition~\ref{#1}}
\newrefformat{eq}{equation~(\ref{#1})}
\newrefformat{app}{Appendix~\ref{#1}}

\usepackage{lscape}

\newenvironment{code-example}
{
\vspace{0.15cm}
\noindent\begin{minipage}{\linewidth}
\begin{center}
\arrayrulecolor{black}
\color{black}
\begin{tabular}{|p{0.95\linewidth}|}
\hline%
\rowcolor{pink!20}%
}
{
\\\hline
\end{tabular}
\end{center}
\end{minipage}
\vspace{-0.2cm}
}

\begin{document}

\title{\vspace{-1ex}\LARGE\textbf{A Survey of Decomposition-Based Evolutionary Multi-Objective Optimization: \textsc{Part I}---Past and Future}}

\author[1]{\normalsize Ke Li}
\affil[1]{\normalsize Department of Computer Science, University of Exeter, EX4 4QF, Exeter, UK}
\affil[\Faxmachine\ ]{\normalsize \texttt{k.li@exeter.ac.uk}}

\date{}
\maketitle

\vspace{-3ex}
{\normalsize\textbf{Abstract: } }Decomposition has been the mainstream approach in classic mathematical programming for multi-objective optimization and multi-criterion decision-making. However, it was not properly studied in the context of evolutionary multi-objective optimization (EMO) until the development of multi-objective evolutionary algorithm based on decomposition (MOEA/D). In this two-part survey series, we use MOEA/D as the representative of decomposition-based EMO to review the up-to-date development in this area, and systematically and comprehensively analyze its research landscape. In the first part, we present a comprehensive survey of the development of MOEA/D from its origin to the current state-of-the-art approaches. In order to be self-contained, we start with a step-by-step tutorial that aims to help a novice quickly get onto the working mechanism of MOEA/D. Then, selected major developments of MOEA/D are reviewed according to its core design components including weight vector settings, subproblem formulations, selection mechanisms and reproduction operators. Besides, we also overview some selected advanced topics for constraint handling, optimization in dynamic and uncertain environments, computationally expensive objective functions, and preference incorporation. In the final part, we shed some light on emerging directions for future developments.

{\normalsize\textbf{Keywords: } }Multi-objective optimization, decomposition, evolutionary computation.


\begin{quote}
    \lq\lq \textit{Divide each difficulty into as many parts as is feasible and necessary to resolve it.}\rq\rq\ \\ --- \textsl{Ren\'e Descartes}
\end{quote}

\section{Introduction}
\label{sec:introduction}

Multi-objective optimization problems (MOPs) are ubiquitous in real-world scenarios, spanning a broad spectrum of fields such as machine learning~\cite{SenerK18}, data mining~\cite{RibeiroZMHLV14}, material science~\cite{AttiaGJSMLCCPYH20}, networkings~\cite{BillingsleyMLMG20,BillingsleyLMMG20,BillingsleyLMMG21}, machine learning robustness~\cite{WilliamsLM23,WilliamsLM23a,Williams023,WilliamsL23b,WilliamsLM23b,Williams0M22,Zhou0M22}, cheminformatics~\cite{LiHWZZL20}, aerospace~\cite{HornbyLL11}, software engineering~\cite{ChenLBY18,LiXCT20,LiX0WT20,Liu0020,ZhouHSL24}, finance and economics~\cite{PonsichJC13}. These problems often involve conflicting objectives, where improving one can compromise others, leading to a set of trade-off solutions known as Pareto-optimal solutions, rather than a single global optimum. Partially due to the implicit parallelism and the population-based nature of evolutionary computation (EC), it has been widely recognized as a major approach for multi-objective optimization (MO). Over the past three decades, significant efforts have been dedicated to developing evolutionary multi-objective optimization (EMO) algorithms. These can be broadly classified into dominance-, indicator-, and decomposition-based frameworks. Representative algorithms include the fast non-dominated sorting genetic algorithm (\textsc{NSGA-II})~\cite{DebAPM02}, the indicator-based evolutionary algorithm (\textsc{IBEA})~\cite{ZitzlerK04}, and the multi-objective evolutionary algorithm based on decomposition (\textsc{MOEA/D})~\cite{ZhangL07}.

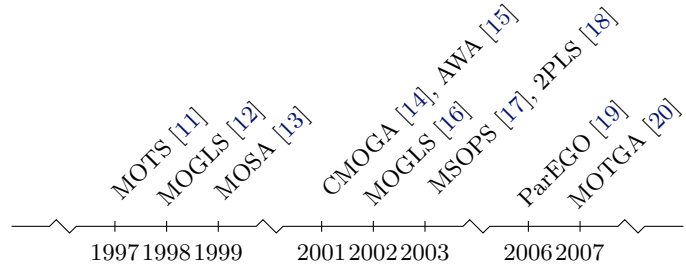
\begin{figure}[t!]
    \centering
    \begin{tikzpicture}[snake=zigzag, scale=0.68, line before snake = 5mm, line after snake = 5mm,font=\footnotesize]
        \draw[snake] (-2,0) -- (0,0);   
        \draw (0,0) -- (2,0);
        \draw[snake] (2,0) -- (4,0);
        \draw (4,0) -- (6,0);
        \draw[snake] (6,0) -- (8,0);
        \draw (8,0) -- (9,0);
        \draw[snake] (9,0) -- (11,0);

        \foreach \x in {0,1,2,4,5,6,8,9}
        \draw (\x cm,3pt) -- (\x cm,-3pt);

        \draw (-2,0) node[below=3pt] {$  $} node[above=3pt] {$  $};
        \draw (-1,0) node[below=3pt] {$  $} node[above=3pt] {$  $};
        \draw (0,0) node[below=3pt] (first) {$1997$} node[above=20pt, rotate=45, right of=first, xshift=-1ex] {MOTS~\cite{Hansen97}};
        \draw (1,0) node[below=3pt] (second) {$1998$} node[above=20pt, rotate=45, right of=second, xshift=-1ex] {MOGLS~\cite{IshibuchiM98}};
        \draw (2,0) node[below=3pt] (third) {$1999$} node[above=20pt, rotate=45, right of=third, xshift=-1ex] {MOSA~\cite{UlunguTFT99}};
        \draw (3,0) node[below=3pt] {$  $} node[above=3pt] {$  $};
        \draw (4,0) node[below=3pt] (fourth) {$2001$} node[above=20pt, rotate=45, right of=fourth, xshift=5ex] {CMOGA~\cite{MurataIG01}, AWA~\cite{JinOS01}};
        \draw (5,0) node[below=3pt] (fifth) {$2002$} node[above=20pt, rotate=45, right of=fifth, xshift=-1ex] {MOGLS~\cite{Jaszkiewicz02b}};
        \draw (6,0) node[below=3pt] (sixth) {$2003$} node[above=20pt, rotate=45, right of=sixth, xshift=5ex] {MSOPS~\cite{PaqueteS03}, 2PLS~\cite{Hughes05}};
        \draw (7,0) node[below=3pt] {$  $} node[above=3pt] {$  $};
        \draw (8,0) node[below=3pt] (seventh) {$2006$} node[above=20pt, rotate=45, right of=seventh, xshift=-1ex] {ParEGO~\cite{Knowles06}};
        \draw (9,0) node[below=3pt] (eighth) {$2007$} node[above=20pt, rotate=45, right of=eighth, xshift=-1ex] {MOTGA~\cite{AlvesA07}};
        \draw (10,0) node[below=3pt] {$  $} node[above=3pt] {$  $};
    \end{tikzpicture}
    \caption{A pragmatic history of decomposition-based EMO algorithms from the past till the development of MOEA/D.}
    \label{fig:timeline}
\end{figure}

Different from the dominance- and indicator-based frameworks, which are rooted in the EC community, decomposition has been the mainstream in the classic mathematical programming and operations research for MO and multi-criterion decision-making (MCDM)~\cite{Miettinen99}. Its core concept involves aggregating different objectives into a scalarizing function by using a dedicated weight vector, transforming a MOP into a single-objective optimization subproblem. The optimum of this subproblem is a Pareto-optimal solution of the original MOP. Traditional mathematical programming approaches, however, heavily rely on derivative information, which is often notoriously hardly accessible in real-world black-box scenarios. Moreover, in MO, these methods can only obtain one Pareto-optimal solution at a time, limiting the effectiveness and flexibility in exploring various trade-off alternatives simultaneously. To the best of our knowledge, MOTS~\cite{Hansen97} is the first that combines the decomposition concept in MCDM with meta-heuristics, tabu search in particular, for MO. As the chronological line shown in~\pref{fig:timeline}, several follow-up works~\cite{IshibuchiM98,UlunguTFT99,MurataIG01,Hughes05,AlvesA07} use a set of randomly generated weight vectors to navigate the EMO search process. A slightly different idea is in~\cite{Jaszkiewicz02b,PaqueteS03,Knowles06} where only a single aggregation function is considered at each iteration and the corresponding weight vector is randomly sampled within a canonical simplex. While~\cite{JinOS01} also only considers one subproblem at a time, the setting of the underpinned weight vector is controlled in a predefined manner. In a nutshell, all these earlier algorithms lack a principled way to maintain population diversity due to the random generation of weight vectors, resulting in less effective search process~\cite{GiagkiozisPF13}. Further, they fail to consider collaboration among neighboring subproblems, rendering them less capable of tackling problems with complex properties~\cite{LiZ09}. The decomposition-based EMO is not established, compared to the dominance- and indicator-based counterparts, until the seminal paper of MOEA/D~\cite{ZhangL07}. 


\noindent
\faVolumeUp\ \underline{\textbf{\textsc{Why is this survey necessary?}}} There are three compelling reasons for organizing this survey series.
\begin{enumerate}
    \item As the foremost representative of decomposition-based EMO methods, MOEA/D transcends being merely a dedicated algorithm; it embodies a pragmatic problem-solving paradigm. Its application as a backbone across a wide spectrum of challenging MOPs underscores its significance and versatility. This constitutes our key motivation of using MOEA/D as the central pillar to develop this survey series.

    \item While there have been nine survey-style articles published since 2014, only~\cite{XuXM20} and~\cite{TrivediSSG17} have offered a comprehensive summary of the development of MOEA/D. Note that these surveys have become somewhat outdated, with some of the topics they highlighted now heavily advanced. In contrast, other articles have focused on more restricted aspects: \cite{PinedaHDPSBM14} and~\cite{WangSLMGLM20} concentrated on decomposition methods, particularly subproblem formulation; \cite{MaYLQZ20} explored weight vector adjustment methods within MOEA/D; and \cite{MaYWJZ16,XuXM19,Guo22a} mainly addressed many-objective challenges for MOEA/D.

    \item To the best of our knowledge, this survey series is the first to offer a comprehensive analysis of the MOEA/D research landscape from a data science perspective. We have developed a novel data mining pipeline that elucidates the categorization and interconnections of research topics within MOEA/D, alongside a spatial-temporal evolution of these topics. Further, our analysis of citation and collaboration networks offers unprecedented insights into the research community's structure. Our analysis results not only shed light on the structural dynamics of MOEA/D research but also extend insights applicable across the entire EMO community and beyond.
\end{enumerate}

\noindent
\faUsers\ \underline{\textbf{\textsc{Who are the audiences?}}} This survey presents techniques and analysis in a progressive manner, starting with a gentle tutorial before delving into more in-depth analysis. This structured approach makes it accessible to diverse audiences. Newcomers and early-career researchers in the EMO community will find it a valuable resource for acquiring a high-level understanding of the EMO domain and identifying pressing research questions for their endeavors. Further, experienced EMO experts are provided with insights into opportunities for creating new research frontiers. Further, the analytic results from our data mining offer added value across different parties in the academic community, aiding researchers, journal editors, and funding agencies in identifying promising research topics and projects, selecting appropriate journals for submissions, and guiding strategic development focus.

\noindent
\faSitemap\ \underline{\textbf{\textsc{Outline:}}} This survey series comes in two parts.
\begin{itemize}
    \item\underline{\textsc{Part I}} focuses on methodological development within the MOEA/D framework where~\pref{fig:mindmap} provides an outline of topics covered therein. \pref{sec:background} begins with a pragmatic tutorial, providing background knowledge pertinent to this survey and making it self-contained. Thereafter, major developments regarding weight vector settings, subproblem formulations, selection mechanisms, and reproduction operators are discussed in~\pref{sec:main_review}, according to the core algorithmic components of MOEA/D. Then, \pref{sec:advanced} and~\pref{sec:others} reviews selected advanced topics and future directions, respectively, which are not exclusive to MOEA/D but are relevant to the wider EMO community. Finally, \pref{sec:conclusions} concludes this survey.

\begin{figure*}[t!]
	\centering
    \includegraphics[width=\linewidth]{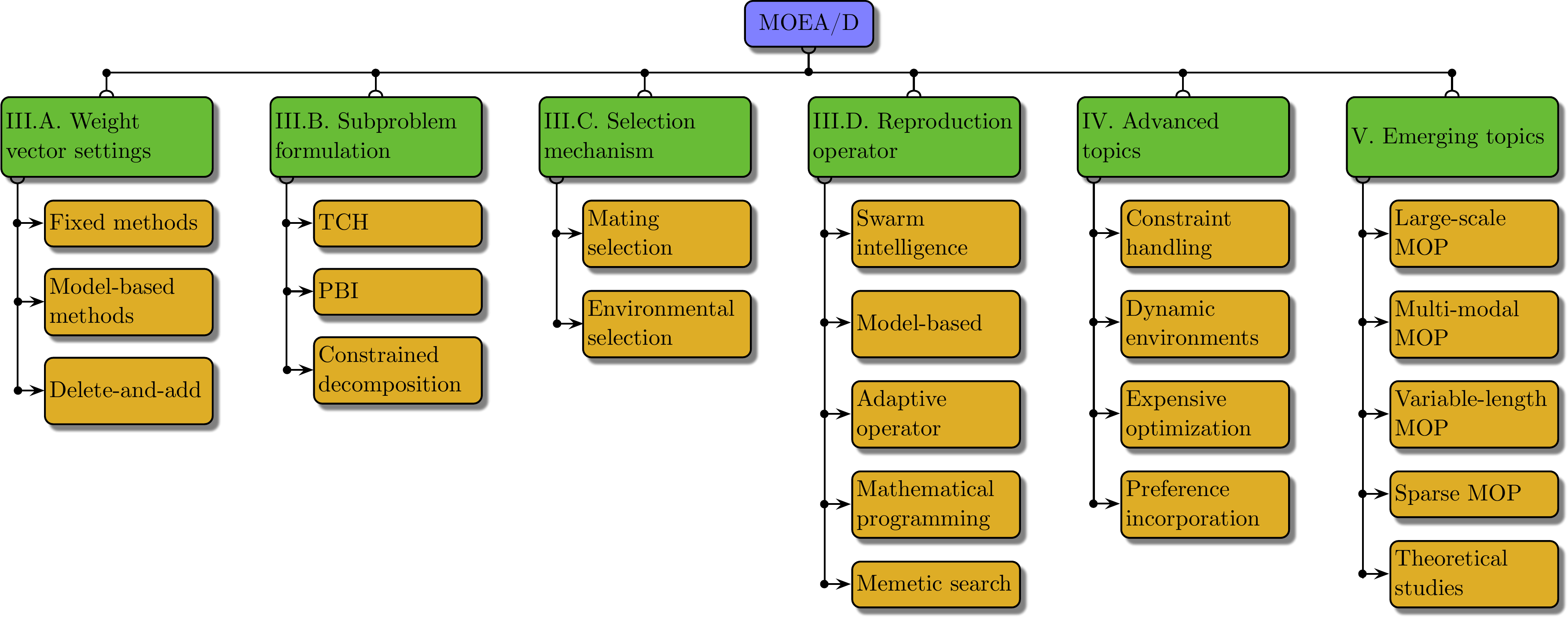}
    \caption{Outline of the topics covered in the \textsc{Part I} of this survey series on decomposition-based EMO.}
    \label{fig:mindmap}
\end{figure*}

    \item\underline{\textsc{Part II}}\footnote{The article of \textsc{Part II} can be downloaded from this \href{https://zenodo.org/records/11032874}{link}, while the \textsc{Appendix} file can be found in this \href{https://zenodo.org/records/11033101}{link}} focuses on the data analytic regarding the current research related to MOEA/D. We will introduce a novel data-driven scientific literature mining platform that acts as the technological foundation to guide data collection, data cleaning, and exploratory data analysis. We believe our data analysis results will offer unprecedented insights into the research community's structure.
\end{itemize}


\section{Background Knowledge}
\label{sec:background}

This section starts with some basic concepts pertinent to this survey, followed by a gentle tutorial on the working mechanisms of a vanilla MOEA/D.

\subsection{Basic Concepts}
\label{sec:concepts}

The MOP\footnote{Note that the definitions herein are interchangeable between minimization and maximization problems.} considered in this paper is defined as:
\begin{equation}
    \begin{array}{l l}
        \underset{\mathbf{x}\in\Omega}{\mathrm{minimize}}\ &\mathbf{F}(\mathbf{x})=(f_{1}(\mathbf{x}),\ldots,f_{m}(\mathbf{x}))^\top\\
        \mathrm{subject\ to} &\mathbf{g}(\mathbf{x})\leq\mathbf{0}\\
        &\mathbf{h}(\mathbf{x})=\mathbf{0}
    \end{array},
    \label{eq:mop}
\end{equation}
where $\Omega=[x_i^\mathrm{L},x_i^\mathrm{U}]^n_{i=1}\subseteq\mathbb{R}^n$ defines the search (or decision variable) space, and $\mathbf{x}=(x_1,\ldots,x_n)^\top$ is a candidate solution therein. $\mathbf{F}:\Omega\rightarrow\mathbb{R}^m$ constitutes $m$ conflicting objective functions, and $\mathbb{R}^m$ is the objective space. $\mathbf{g}(\mathbf{x})=(g_1(\mathbf{x}),\ldots,g_p(\mathbf{x}))^\top$ and $\mathbf{h}(\mathbf{x})=(h_1(\mathbf{x}),\ldots,h_q(\mathbf{x}))^\top$ consists of $p\ge 1$ inequality and $q\ge 1$ equality constraints, respectively. A solution $\mathbf{x}\in\Omega$ is \underline{\textit{feasible}} if and only if $\mathbf{g}(\mathbf{x})\leq\mathbf{0}$ and $\mathbf{h}(\mathbf{x})=\mathbf{0}$.

\begin{definition}
    Given two feasible solutions $\mathbf{x}^1$ and $\mathbf{x}^2$, $\mathbf{x}^1$ is said to \underline{\textit{Pareto dominate}} $\mathbf{x}^2$ (denoted as $\mathbf{x}^1\preceq\mathbf{x}^2$) if and only if $f_i(\mathbf{x}^1)\leq f_i(\mathbf{x}^2)$, $\forall i\in\{1,\ldots,m\}$ and $\exists j\in\{1,\ldots,m\}$ such that $f_i(\mathbf{x}^1)<f_i(\mathbf{x}^2)$.
\end{definition}

\begin{definition}
    A feasible solution $\mathbf{x}^\ast$ is \underline{\textit{Pareto-optimal}} with respect to (\ref{eq:mop}) if $\nexists\mathbf{x}\in\Omega$ with $\mathbf{g}(\mathbf{x})\leq\mathbf{0}$ and $\mathbf{h}(\mathbf{x})=\mathbf{0}$ such that $\mathbf{x}\preceq\mathbf{x}^{\ast}$.
\end{definition}

\begin{definition}
    The set of all Pareto-optimal solutions is called the \underline{\textit{Pareto-optimal set}} (PS). Accordingly, $PF=\{\mathbf{F}(\mathbf{x})|\mathbf{x}\in PS\}$ is called the \underline{\textit{Pareto-optimal front}} (PF).
\end{definition}

\begin{definition}
    The \underline{\textit{ideal objective vector}} $\mathbf{z}^{\mathrm{id}}=(z^{\mathrm{id}}_1,\ldots,z^{\mathrm{id}}_m)^\top$ is constructed by the minimum of each objective function, i.e., $z^{\mathrm{id}}_i=\underset{\mathbf{x}\in\Omega}{\min}f_i(\mathbf{x})$, $i\in\{1,\ldots,m\}$.
\end{definition}

\begin{definition}
    The \underline{\textit{nadir objective vector}} $\mathbf{z}^\mathrm{nad}=(z^{\mathrm{nad}}_1,\ldots,z^{\mathrm{nad}}_m)^\top$ is constructed by the worst objective functions of the PF, i.e., $z^{\mathrm{nad}}_i=\underset{\mathbf{x}\in PS}{\max}f_i(\mathbf{x})$, $i\in\{1,\ldots,m\}$.
\end{definition}

\subsection{Tutorial of Vanilla MOEA/D}
\label{sec:moead}

Generally speaking, a MOEA/D consists of two major functional components, i.e., \underline{\textit{decomposition}} and \underline{\textit{collaboration}}. Their working mechanisms are briefly introduced as follows.

\subsubsection{Decomposition}
\label{sec:decomposition}

Under some mild continuous conditions, a Pareto-optimal solution of a MOP is an optimal solution of a scalar optimization problem whose objective function is an aggregation of all individual objectives~\cite{Miettinen99}. The principle of decomposition is to transform the task of approximating the PF into a number of scalarized subproblems. There have been various approaches for constructing such subproblems~\cite{Miettinen99}, which will be further discussed in~\pref{sec:subproblems}. Here we introduce the three most widely used ones within the MOEA/D framework. 
\begin{itemize}
    \item\underline{\textit{Weighted sum (WS)}}: it is a convex combination of all individual objectives. Let $\mathbf{w}=(w_1,\ldots,w_m)^\top$ be a weight vector where $w_i\geq 0$, $\forall i\in\{1,\ldots,m\}$ and $\sum_{i=1}^m w_i=1$, the WS is formulated as:
        \begin{equation}
            \underset{\mathbf{x}\in\Omega}{\mathrm{minimize}}\ g^{\mathrm{ws}}(\mathbf{x}|\mathbf{w})=\sum_{i=1}^{m}w_{i}f_i(\mathbf{x}).
            \label{equation:WS}
        \end{equation}

    \item\underline{\textit{Weighted Tchebycheff (TCH)}}: in this approach, the scalar optimization problem is formulated as:
        \begin{equation}
            \underset{\mathbf{x}\in\Omega}{\mathrm{minimize}}\ g^{\mathrm{tch}}(\mathbf{x}|\mathbf{w},\mathbf{z}^{\ast})=\max\limits_{1\leq i\leq m}\big\{w_i|f_i(\mathbf{x})-z_{i}^{\ast}|\big\},
            \label{equation:TCH}
        \end{equation}
        where $\mathbf{z}^\ast$ is the approximated ideal objective vector estimated by using the current population.

    \item\underline{\textit{Penalty-baed boundary intersection (PBI)}}: this is a variant of the normal-boundary intersection method~\cite{DasD98}, whose equality constraint is handled by a penalty function. Formally, it is defined as:
        \begin{equation}
            \underset{\mathbf{x}\in\Omega}{\mathrm{minimize}}\ g^{\mathrm{pbi}}(\mathbf{x}|\mathbf{w},\mathbf{z}^{\ast})=d_1+\theta d_2,
            \label{eq:pbi}
        \end{equation}
        where $\theta>0$ is a user-defined penalty parameter, $d_1=\|(\mathbf{F}(\mathbf{x})-\mathbf{z}^{\ast})^\top\mathbf{w}\|/\|\mathbf{w}\|$ and $d_2=\|\mathbf{F}(\mathbf{x})-(\mathbf{z}^{\ast}+d_1\mathbf{w})\|$.
\end{itemize}

Note that MOEA/D applies the Das and Dennis's method~\cite{DasD98} to specify a set of weight vectors evenly distributed along a unit simplex. Its basic idea is to divide each coordinate into $H>0$ equally spaced segments. The weight vectors are constituted by picking up a sliced coordinate along each axis in an iterative manner. In total, this method can generate ${H+m-1}\choose{m-1}$ weight vectors, each of which corresponds to a unique subproblem in MOEA/D. Fig. A1 of the \textsc{Appendix} gives an example that illustrates $21$ weight vectors generated in a three-dimensional space when $H=5$.

\begin{figure*}[t!]
	\centering
    \includegraphics[width=\linewidth]{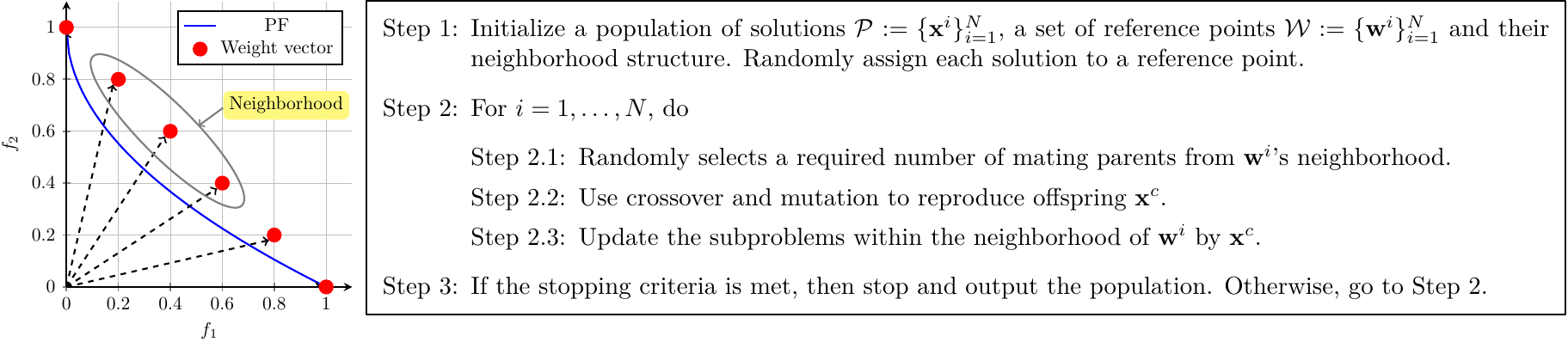}
    \caption{The left panel provides an illustrative example of the weight vector distribution and the neighborhood structure in MOEA/D. The right panel gives the working mechanism of MOEA/D as a three-step process.}
    \label{fig:moead}
\end{figure*}

\subsubsection{Collaboration}
\label{sec:collaboration}

Different from the classic mathematical programming method for MO, in which only one subproblem is considered at a time, MOEA/D applies a population-based meta-heuristic approach to concurrently solve all subproblems in a collaborative manner. Each subproblem is associated with its best-so-far solution during the evolutionary problem-solving process. The basic assumption behind this collaboration mechanism is that neighboring subproblems are more likely to share similar properties, e.g., similar objective functions and/or optimal solutions. In particular, the neighborhood of a subproblem is determined by the Euclidean distance of its corresponding weight vector with respect to the others. By this means, solutions associated with the suproblems lying in the same neighborhood can collaborate with each other to push the evolution forward by sharing and exchanging elite information. Specifically, the mating parents are picked up from the underlying neighborhood (as depicted in the left panel of~\pref{fig:moead}) for offspring reproduction while the newly generated offspring is used to update the subproblem(s) within the same neighborhood. Note that the solution associated with a subproblem can be updated/replaced by the newly generated offspring if and only if it has a better fitness value.

As in the right panel of~\pref{fig:moead}, the skeleton of MOEA/D consists of four algorithmic components, including the setting of weight vectors, the subproblem formulation, the mating or environmental selection, and the offspring reproduction. In the next section, our discussion is mainly guided by the primary studies dedicated to one or multiple pieces of these above algorithmic components within the MOEA/D framework.



\section{Selected Developments on Major Components of MOEA/D}
\label{sec:main_review}

In this section, we review some selected developments in the four major components of MOEA/D.


\subsection{Weight Vector Setting}
\label{sec:weight}

\begin{figure*}[htbp]
	\centering
    \includegraphics[width=\linewidth]{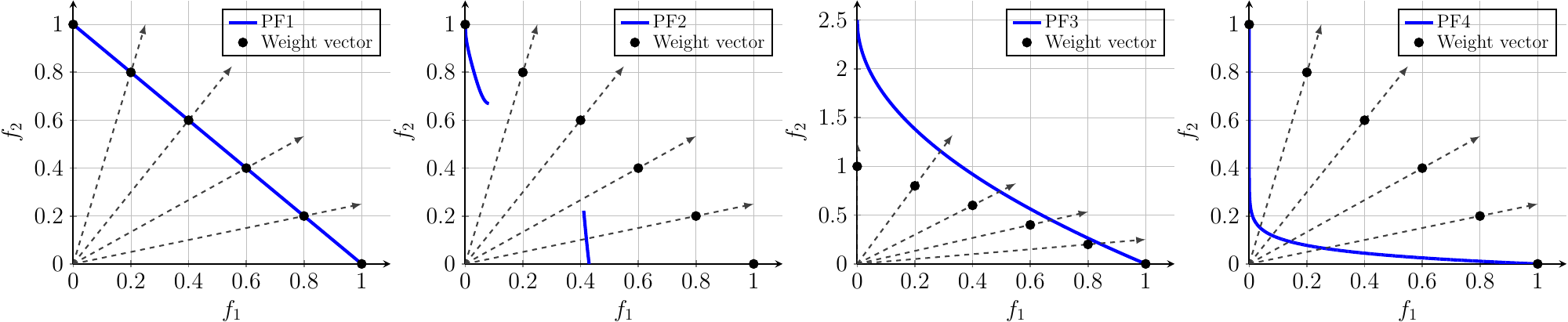}
    \caption{MOEA/D obtains evenly distributed solutions by using evenly distributed weight vectors merely when the PF is a simplex. Otherwise, some weight vectors do not have Pareto-optimal solutions like PF2; or the distribution of the corresponding Pareto-optimal solutions is highly biased like PF3 and PF4.}
    \label{fig:weight_discussion}
\end{figure*}

In MOEA/D, weight vectors act as carriers for subproblems, with their configuration being crucial for achieving a well-balanced convergence and diversity~\cite{IshibuchiSMN17}. However, employing evenly distributed weight vectors may prove detrimental for PFs that are not regular simplexes, exhibiting disconnected, badly scaled, or strongly convex shapes, as illustrated in~\pref{fig:weight_discussion}. To address this issue, a significant body of work has focused on proactively adapting weight vectors. This subsection reviews selected developments in this area, categorized by the type of heuristics used for adaptation, as summarized in Table A1 of the \textsc{Appendix}.

\subsubsection{Fixed methods}
\label{sec:fixed_method}

Rather than progressively tuning weight vectors, this approach employs a fixed strategy to set the weight vectors based on certain regularities \textit{a priori}. The simplest way is random sampling along the canonical simplex. For instance, MOGLS~\cite{IshibuchiM98,Jaszkiewicz02a,Jaszkiewicz02b} guides a local search on an elite solution with a randomly generated weight vector for each generation. In~\cite{LiDDZ15}, evenly distributed weight vectors are switched to a random sampling when the search stagnates.

Beyond pure randomness, principled methods from optimal experimental design, such as uniform design~\cite{TanJLW12,TanJLW13,TanJ13,GaoNL19}, have been utilized for weight vector generation. Deb and Li et al.~\cite{DebJ14,LiDZK15} introduced a two-layer method to enhance weight vector distribution within the canonical simplex for many-objective optimization. Jin et al.~\cite{JinOS01} proposed a periodical variation method for two-objective problems, where a weight vector $\mathbf{w}(t)=(w_1(t),w_2(t))^\top$ changes gradually with the evolution process as ${w_1(t)=|\sin(2\pi t)/F|,w_2(t)=1.0-w_1(t)}$, with $F$ controlling the change frequency. In~\cite{BlankDDBS21}, the Riesz $s$-energy metric was used to iteratively generate well-spaced weight vectors in high-dimensional spaces.

\subsubsection{Model-based adaptation methods}
\label{sec:model_method}

The basic concept behind model-based adaptation methods in MOEA/D is to proactively estimate the PF shape by learning from the evolutionary population or an archive, and then resample weight vectors from this estimated PF. These methods fall into two main categories based on their modeling approach.

The first research direction changes the PF assumption from a simplex format, $\sum_{i=1}^m f_i(\mathbf{x})=1$, to a parameterized polynomial format, $\sum_{i=1}^m f^p_i(\mathbf{x})=1$. Practical implementations estimate the exponential term $p$ either by solving a nonlinear least square problem~\cite{LiuGC10} or by maximizing Hypervolume~\cite{JiangCZO11}. However, this approach struggles with complex PF shapes, thus~\cite{JiangZO11} proposed an asymmetric Pareto-adaptive scheme using a curve model for two-objective PFs. DMOEA/D~\cite{GuLT12} addresses disconnected PF segments with piecewise hyperplanes estimated from non-dominated solutions.

In contrast to fixed assumptions, another research strand employs parametric models learned from data, such as the evolutionary population. For example, in~\cite{WuKJLZ17} and~\cite{WuLKZZ19}, Gaussian process regression was employed to model the PF under a general weighted polynomial assumption. This approach involves initially sampling a large number of weight vectors from the model, then refining the sample set by removing dominated or high-variance predictions. This method effectively filters out samples in disconnected regions or beyond the PF. More recently, Wang et al.~\cite{WangGCGX23} proposed a cooperative strategy to alternate between global and local weight vectors, thus improving the distribution and uniformity of the population on complicated bi-objective optimization problems. MOEA/D-SOM uses the self-organizing map (SOM)~\cite{Kohonen82} to learn a latent model of the PF and the topological properties are preserved through the use of neighborhood functions~\cite{GuC18}. A set of uniformly distributed weight vectors is thereafter sampled from the SOM model. Later, the growing neural gas model~\cite{Fritzke94}, a variant of SOM, is applied in a similar manner to facilitate the weight vector generation in many-objective optimization~\cite{LiuIMN20,LiuJHRY22}. Wang et al.~\cite{WangHP23} proposed using a supervised self-organizing network LVQ to model the distribution of weight vectors in the feasible region(s).

\subsubsection{Delete-and-add methods}
\label{sec:delete_and_add_method}

They focus on local tuning by removing invalid weight vectors---those in regions without feasible solutions or overly crowded areas---and adding new ones in promising regions. A representative approach is MOEA/D-AWA~\cite{QiMLJSW14}, which periodically deletes weight vectors from crowded regions, guided by an external archive, and introduces new ones in sparser areas. This strategy aims to improve exploration in many-objective optimization problems by initially generating weight vectors randomly~\cite{FariasBBA18,FariasA19} and focusing on under-exploited areas~\cite{LiY20,WangWW16}.

Contrary to targeting sparse regions, some methods generate new weight vectors away from current promising areas. RVEA$^\ast$, for example, partitions the population and replaces associated weight vectors of empty subpopulations with unit vectors generated within specific ranges~\cite{ChengJOS16}. This approach aims to diversify exploration by introducing variability in the weight vector distribution. Another strategy involves generating new weight vectors within a regulated neighborhood, such as creating a simplex with neighboring weight vectors for each crowded one~\cite{JainD14,JainD13,CamachoPBI19}. This method seeks to maintain a balance between exploration and exploitation by focusing on less crowded but potentially promising areas. Some methods reset weight vectors based on the distribution of non-dominated solutions within local niches, directly using archive solutions to adjust weight vectors~\cite{GuoWW15,JiangFHNOZT16,TianCZCJ18}. It enhances the algorithm's ability to adapt to the evolving distribution of solutions, aiming for a more targeted exploration in the objective space.

\begin{tcolorbox}[breakable, title after break=, height fixed for = none, colback = gray!40!white, boxrule = 0pt, sharpish corners, top = 4pt, bottom = 4pt, left = 4pt, right = 4pt, toptitle=2pt, bottomtitle=2pt]
    \faBomb\ Key bottlenecks of the current weight vector adaptation methods include: $\blacktriangleright$ Fixed methods are notorious for being too rigid to accommodate versatile PF shapes in the real world. $\blacktriangleright$ While both model-based adaptation and delete-and-add methods are designed to adapt to the PF shape on the fly, their success heavily depends on the effectiveness of the evolutionary process itself. There is a risk that the weight vector adaptation may be compromised by the populations trapped in local optima or deceptive regions of the search space. $\blacktriangleright$ Due to curse-of-dimensionality, representing a high-dimensional space with a limited set of weight vectors is highly challenging, while evolving with a huge number of weight vectors significantly increases computational demands.
\end{tcolorbox}


\subsection{Subproblem Formulations}
\label{sec:subproblems}

Subproblem formulation determines the search behavior of MOEA/D. Most subproblem formulations used in MOEA/D are derived from the three traditional ones introduced in~\pref{sec:decomposition}. In this paper, our focus, as the selected works summarized in Table A2 of the \textsc{Appendix}, is on the shape of the contour line and the size of the improvement region, which defines areas with superior solutions. Specifically, a larger improvement region indicates stronger selection pressure towards convergence, whereas a smaller one emphasizes diversity.

\subsubsection{TCH variants}
\label{sec:tch_variants}

\begin{figure*}[htbp]
	\centering
    \includegraphics[width=.85\linewidth]{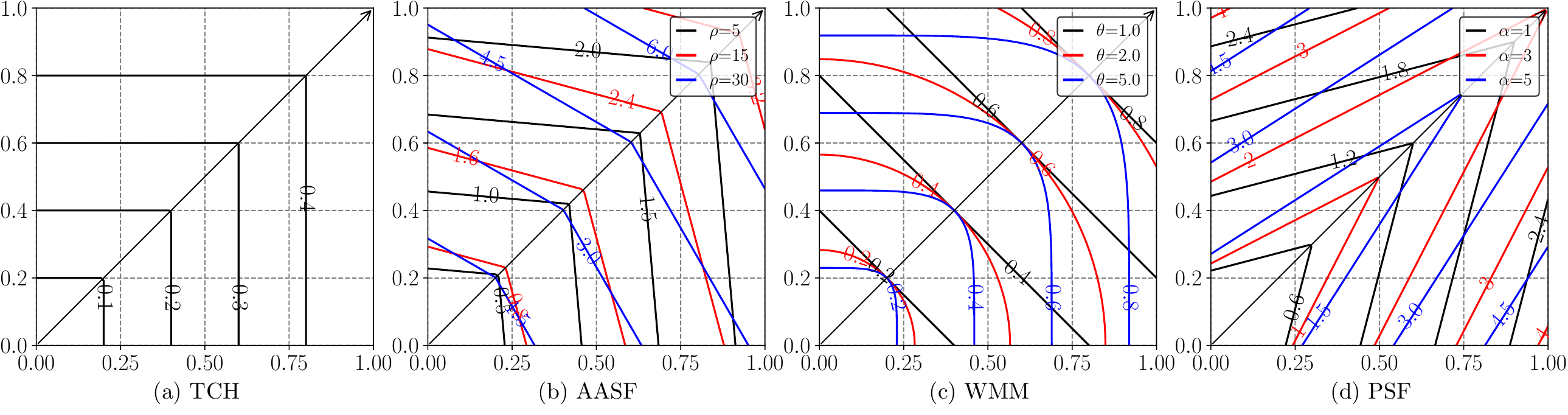}
    \caption{Contour lines of the Tchebycheff approach and its three variants.}
    \label{fig:contour_tch}
\end{figure*}

The shape of the contour line in the TCH approach and its variants primarily depends on the distance measure relative to the ideal point. As illustrated in~\pref{fig:contour_tch}(a), the Tchebycheff distance, utilized in the original TCH approach, generates a contour line that mirrors the Pareto dominance relation. This resemblance underpins the TCH approach's asymptotic equivalence to the Pareto dominance relation in identifying superior solutions~\cite{GiagkiozisF15}. Despite the TCH approach not being limited by the convexity of the PF, unlike the WS approach, it struggles to distinguish between weakly dominated solutions. A direct alternative to overcome this limitation is the implementation of an augmented achievement scalarizing function (AASF)~\cite{Miettinen99}, defined as:
\begin{align}
    \min_{\mathbf{x}\in\Omega}\quad g^{\mathrm{aasf}}(\mathbf{x}|\mathbf{w},\mathbf{z}^\ast)&=\max_{1\leq i\leq m}\bigg\{\frac{f_i(\mathbf{x})-z_i^\ast}{w_i}\bigg\}\nonumber\\ &+\rho\sum_{i=1}^m\bigg\{\frac{f_i(\mathbf{x})-z_i^\ast}{w_i}\bigg\},
    \label{eq:asf}
\end{align}
where $\rho>0$ is a parameter that adjusts the opening angle of the contour line, as demonstrated in Section\pref{fig:contour_tch}(b). However, setting the $\rho$ value in practice to control the improvement region lacks a thumb rule. Another notable application of the AASF is found in~\cite{WuLKZ20}, where an adversarial decomposition approach is proposed to capitalize on the complementary characteristics of the AASF and conventional PBI approaches.

Another critique of the TCH approach concerns its crisp improvement region, which can lead to a loss of population diversity. To address this problem, a natural idea is to adapt the shape of the contour line to each local niche of the corresponding subproblem. For instance, Wang et al.~\cite{WangZZ15,WangZZ16} explored the characteristics of a family of weighted metrics methods within the MCDM community, defined as follows:
\begin{equation}
    \min_{\mathbf{x}\in\Omega}\quad g^{\mathrm{wmm}}(\mathbf{x}|\mathbf{w},\mathbf{z}^\ast)=\bigg(\sum_{i=1}^m w_i|f_i(\mathbf{x})-z_i^\ast|^\theta\bigg)^{\frac{1}{\theta}},
\end{equation}
where $\theta\geq 1$ controls the geometric characteristics of the contour lines, as depicted in~\pref{fig:contour_tch}(c). By varying $\theta$ within a predefined set, the improvement region can be fine-tuned for each subproblem. Expanding on this concept, Jiang et al.~\cite{JiangYWL18} introduced a multiplicative scalarizing function (MSF), further refining the adjustment of contour shapes to balance convergence and diversity:
\begin{equation}
    \min_{\mathbf{x}\in\Omega}\quad g^{\mathrm{msf}}(\mathbf{x}|\mathbf{w},\mathbf{z}^\ast)=\frac{[\max_{1\leq i\leq m}(\frac{1}{w_i}|f_i(\mathbf{x})-z_i^\ast|)]^{1+\alpha}}{[\min_{1\leq i\leq m}(\frac{1}{w_i}|f_i(\mathbf{x})-z_i^\ast|)]^{\alpha}},
\end{equation}
where $\alpha$ is a parameter that controls the shape of the contours of the MSF, which becomes the TCH when $\alpha=0$ while it overlaps with the corresponding weight vector when $\alpha=+\infty$. To have an adaptive control of $\alpha$,~\cite{JiangYWL18} suggested that:
\begin{equation}
    \alpha^i=\beta(1-g/G_{\max})\{m\times\min_{1<j<m}(w_j^i)\},
\end{equation}
where $\alpha^i$ is the $\alpha$ value associated with the $i$-th subproblem. Meanwhile, the authors developed another TCH variant called penalty-based scalarizing function (PSF), the contour lines of which are given in~\pref{fig:contour_tch}(d), and it is defined as:
\begin{equation}
    \min_{\mathbf{x}\in\Omega}\quad g^{\mathrm{psf}}(\mathbf{x}|\mathbf{w},\mathbf{z}^\ast)=\max_{1\leq i\leq m}(\frac{1}{w_i}|f_i(\mathbf{x})-z_i^\ast|)+\alpha d,
\end{equation}
where $d$ is the same as $d_2$ in the PBI approach and $\alpha$ is a parameter that controls the balance between convergence and diversity. In particular, a larger $\alpha$ value prioritizes the diversity. In practice, the authors proposed to linearly decrease $\alpha$ with the evolution progress and it becomes zero at the end. This makes both MSF and PSF gradually degenerate to TCH. Thus, the corresponding improvement region for each subproblem is accordingly enlarged leading to a more emphasis on the convergence and relaxation of the diversity.

\begin{figure*}[htbp]
	\centering
    \includegraphics[width=.85\linewidth]{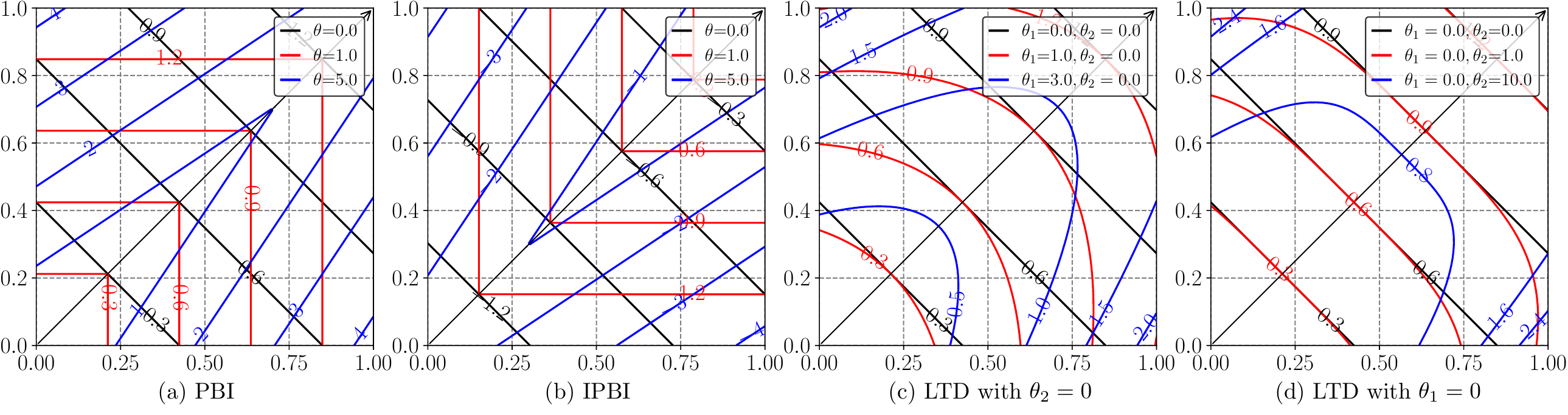}
    \caption{Contour lines of the PBI approach and its three variants.}
    \label{fig:contour_pbi}
\end{figure*}

\subsubsection{PBI variants}
\label{sec:pbi_variants}

As a parameterized combination of two distance measures, the improvement region of the PBI approach is controlled by $\theta$ in~\pref{eq:pbi}. As shown in~\pref{fig:contour_pbi}(a), it becomes the WS approach when $\theta=0.0$ and the TCH approach when $\theta=1.0$. Its improvement region becomes smaller with the increase of $\theta$. One typical idea of the PBI variants is largely about the adaptation of $\theta$ with respect to the shape of the PF~\cite{YangJJ17,MingWZZ17}. For example, Yang et al.~\cite{YangJJ17} proposed two $\theta$ adaptation methods. One is called the adaptive penalty scheme (APS) that gradually increases the $\theta$ value with the progression of evolution: 
\begin{equation}
\theta=\theta_{\min}+(\theta_{\max}-\theta_{\min})\frac{t}{t_{\max}},
\end{equation}
where $t$ is the generation count and $t_{\max}$ is the maximum number of generations. $\theta_{\min}=1.0$ and $\theta_{\max}=10.0$ are the lower and upper boundary of $\theta$, respectively. The basic idea of APS is to promote convergence during the early stage of evolution while gradually shifting to diversity. The alternative way, called a subproblem-based penalty scheme (SPS), is to assign an independent $\theta$ value for each subproblem:
\begin{equation}
\theta_i=e^{ab_i}, b_i=\max_{1\leq j\leq m}\big\{w_i^j\big\}-\min_{1\leq j\leq m}\{w_i^j\},
\end{equation}
where $\theta_i$ means a $\theta$ value for the $i$-th subproblem in PBI and $w_i^j$ is the $j$-th element for the $i$-th weight vector. $a>0$ is a scaling factor that controls the magnitude of the penalty. 

Instead of tweaking the penalty term in the PBI approach, Sato proposed an inverted PBI (IPBI)~\cite{Sato14,Sato15}:
\begin{equation}
	\min_{\mathbf{x}\in\Omega}\quad g^{\mathrm{ipbi}}(\mathbf{x}|\mathbf{w},\mathbf{z}^{\mathrm{nad}})=d_1^{\mathrm{nad}}-\theta d_2^{\mathrm{nad}},
\end{equation}
where $d_1^{\mathrm{nad}}=\|\mathbf{z}^{\mathrm{nad}}-\mathbf{F}(\mathbf{x})\mathbf{w}\|/\|\mathbf{w}\|$ and $d_2^{\mathrm{nad}}=\big\|\mathbf{z}^{\mathrm{nad}}-\big(\mathbf{F}(\mathbf{x}+d^{\mathrm{nad}}_1\frac{\mathbf{w}}{\|\mathbf{w}\|}\big)\big\|$. Its basic idea is to push the solution away from $\mathbf{z}^{\mathrm{nad}}$ as much as possible. From the examples shown in~\pref{fig:contour_pbi}(b), we find that the IPBI approach essentially shares similar characteristics as the conventional PBI approach but it can search a wider range of the objective space. In~\cite{WuLKZZ19}, Wu et al. proposed an augmented format of the PBI approach as:
\begin{equation}
	\min_{\mathbf{x}\in\Omega}\; y(\mathbf{x}|\mathbf{n}^\ast,\mathbf{z}^\ast)=h(\mathbf{\overline{F}(x)}|\mathbf{n}^\ast,\mathbf{z}^\ast)=d_1+\theta_1 d_2^2+\theta_2 d_2^4,
\end{equation}
where $\theta_1>0$ and $\theta_2>0$ are parameters that control the shape and distribution of the opening angle and curvature contours and thus search behaviors of the underlying subproblem as shown in~\pref{fig:contour_pbi}. In practice, $\theta_1$ and $\theta_2$ are set according to the manifold structure of the underlying PF estimated by a Gaussian process regression model.

\begin{figure*}[htbp]
	\centering
    \includegraphics[width=\linewidth]{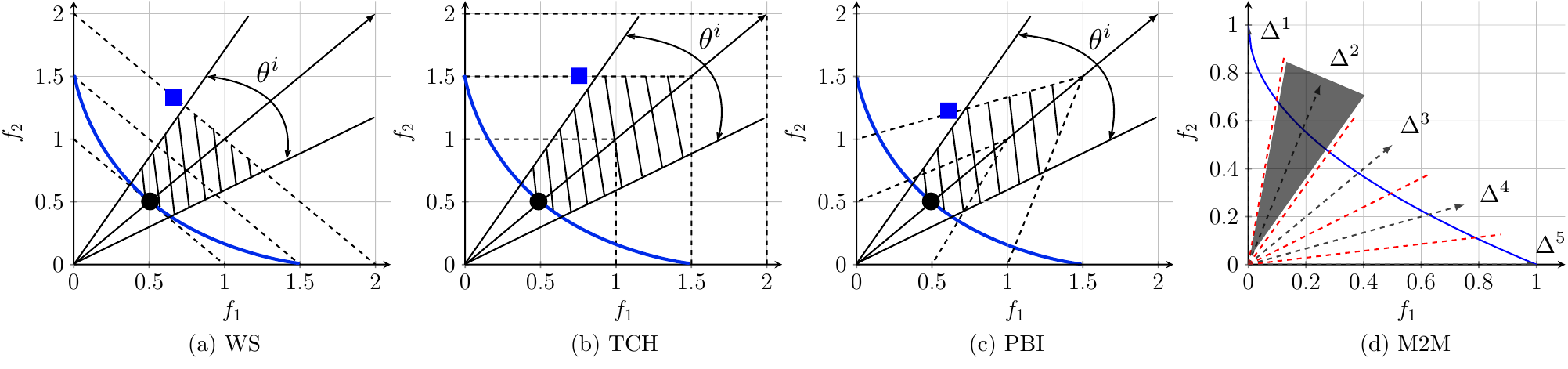}
    \caption{Contour lines of different variants CD (black circle is not comparable with the blue square due to the constrained angle $\theta^i$).}
    \label{fig:contour_cd}
\end{figure*}

\subsubsection{Constrained Decomposition}
\label{sec:constrained_decomposition}

Instead of working on the entire search space, constrained decomposition (CD) is to restrict the improvement region of each subproblem to be a local niche~\cite{WangZZGJ16}. This idea is similar to some TCH and PBI variants (e.g.,~\cite{WangZZ15,WangZZ16,MingWZZ17,YangJJ17,WuLKZZ19}) that applied a dedicated parameterization on each subproblem. Specifically, a constrained optimization subproblem in CD is defined as:
\begin{equation}
\begin{array}{l l}
\min\quad\quad\quad\; g(\mathbf{x}|\mathbf{w}^i,\mathbf{z}^\ast)\\
\mathrm{subject\ to} \quad \mathbf{x}\in\Omega,\langle\mathbf{w}^i,\mathbf{F}(\mathbf{x})-\mathbf{z}^\ast \rangle\leq 0.5\theta^i
\end{array},
\end{equation}
where $g(\cdot)$ can be any subproblem formulation and $\theta^i$ controls the improvement region of the $i$-th subproblem. Based on this definition, the one that meets the constraint is superior to the violated one(s). If both solutions meet the constraint or not, they are compared based on the $g(\cdot)$ function. As depicted in~\pref{fig:contour_cd}(a) to (c), the improvement region of the constrained subproblem is much smaller than its conventional version, enabling a better population diversity. In~\cite{WangZILZ18}, a local constraint is imposed for WS to come up with a localized WS:
\begin{equation}
\begin{array}{l l}
\min\quad\quad\quad\;  g(\mathbf{x}|\mathbf{w}^i,\mathbf{z}^\ast)\\
\mathrm{subject\ to} \quad \mathbf{x}\in\Omega,\langle\mathbf{w}^i,\mathbf{F}(\mathbf{x})\rangle\leq\omega^i
\end{array},
\end{equation}
where $\omega^i=\sum_{j=1}^m\frac{\theta^{ij}}{m}$ is the constraint for the $i$-th subproblem and $\theta^{ij}$ is an acute angle between the $i$-th weight vector and its $j$-th closest weight vector. In particular, $g(\mathbf{x}^\ast|\mathbf{w}^i,\mathbf{z}^\ast)$ is set to infinity when $\langle\mathbf{w}^i,\mathbf{F}(\mathbf{x}^\ast)\rangle>\omega^i$, i.e., $\mathbf{x}^\ast$ is out of comparison when consider the $\omega^i$ constraint. Each constraint imposes a hypercone region regarding each subproblem, similar to the effect of the example in~\pref{fig:contour_cd}(a). By doing so, we can expect to make solution(s) lying in the non-convex region survive in the update procedure when using the localized WS approach.

Instead of decomposing the original MOP into a scalar optimization problem, Liu et al.~\cite{LiuGZ14} proposed MOEA/D-M2M that decomposes a MOP into $K>1$ simplified MOPs where the $i$-th one is defined as:
\begin{equation}
\begin{array}{l l}
	\min\quad\quad\quad \mathbf{F}(\mathbf{x})=(f_1(\mathbf{x}),\ldots,f_m(\mathbf{x}))^\top\\
\mathrm{subject\ to}\;\;\; \mathbf{x}\in\Omega,\mathbf{F}(\mathbf{x})\in\Delta^i\\
\end{array},
\end{equation}
where $\Delta^i=\{\mathbf{F}(\mathbf{x})\in\mathbb{R}^m|\langle\mathbf{F}(\mathbf{x}),\mathbf{w}^i\rangle\leq\langle\mathbf{F}(\mathbf{x}),\mathbf{w}^j\rangle\}$, $j\in\{1,\ldots,N\}$ and $\langle\mathbf{F}(\mathbf{x}),\mathbf{w}\rangle$ is the acute angle between $\mathbf{F}(\mathbf{x})$ and $\mathbf{w}$. \pref{fig:contour_cd}(d) provides an illustrative example of five simplified MOPs in a two-objective space. In practice, each simplified MOP is solved by NSGA-II collaboratively. Since each subregion is enforced with a population of solutions, MOEA/D-M2M naturally strikes a balance between convergence and diversity. The effectiveness of this M2M decomposition approach is validated on problems with some challenging bias towards certain regions of the PF as well as those with disconnected segments~\cite{LiuCDG17,LiuCZD18}.

\begin{tcolorbox}[breakable, title after break=, height fixed for = none, colback = gray!40!white, boxrule = 0pt, sharpish corners, top = 4pt, bottom = 4pt, left = 4pt, right = 4pt, toptitle=2pt, bottomtitle=2pt]
    \faBomb\ Key bottlenecks of the existing subproblem formulations include: $\blacktriangleright$ All existing subproblem formulations within MOEA/D are parametric. Thereby, their search behaviors are influenced by specific parameter settings. However, there is no rule of thumb for selecting the optimal parameters for black-box problems. $\blacktriangleright$ Attempts to adaptively set parameters based on the current population lead to the risk of biasing the search towards premature convergence. $\blacktriangleright$ Further, the varying difficulties of different subproblems present another challenge---optimally allocating computational resources based on their evolutionary status remains an unresolved issue.
\end{tcolorbox}


\subsection{Selection Mechanisms in MOEA/D}
\label{sec:selection}

Selection in MOEA/D plays a crucial role in two key processes---\textit{mating selection}, which determines the selection of high-quality parents for offspring reproduction, and \textit{environmental selection}, which ensures the survival of the fittest and guides the evolution. Table A$3$ of \textsc{Appendix} summarizes related works on both selection mechanisms.

\begin{figure*}[htbp]
	\centering
    \includegraphics[width=\linewidth]{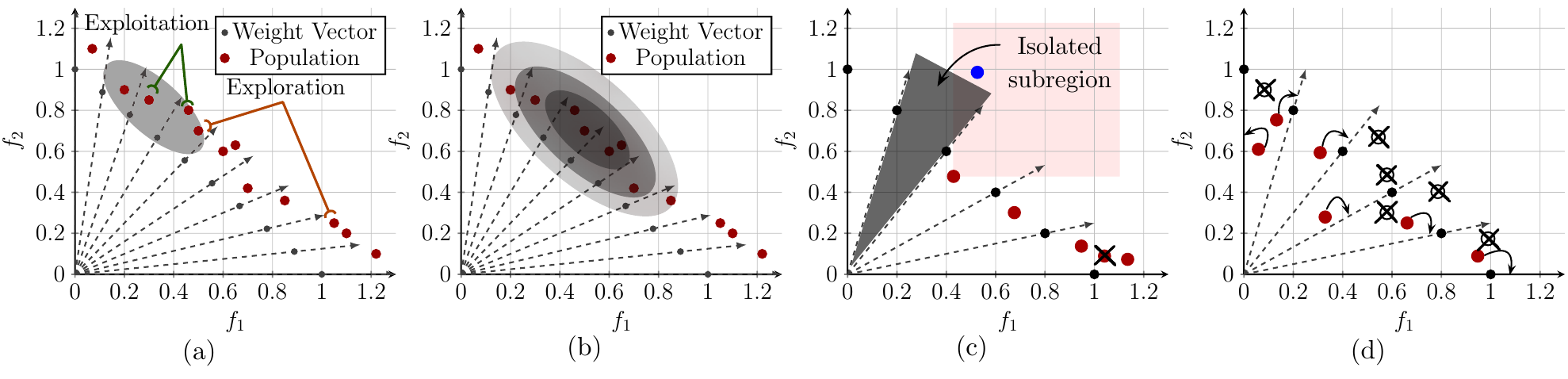}
    \caption{Illustrative examples of selected mating selection mechanisms and environmental selection mechanisms in MOEA/D. (a) green and red lines represent mating selection within and outside the neighborhood, respectively. (b) three different neighborhood sizes are shaded in different colors. (c) the blue point is dominated but lies in an isolated subregion. (d) the eliminated solutions are marked with crosses.}
    \label{fig:selection}
\end{figure*}

\subsubsection{Mating selection}
\label{sec:mating_selection}

In MOEA/D~\cite{ZhangL07}, the mating selection is restricted to the neighborhood of each subproblem. This is based on the assumption that neighboring subproblems share certain structural similarities thus their corresponding solutions are expected to help each other. However, this mechanism is too exploitive, leading to a risk of being trapped by local optima. A simple strategy, as depicted in~\pref{fig:selection}(a), is to allow mating parents to be picked up from the entire population with a low probability~\cite{LiZ09}. Differently, \cite{JiangY16} proposed taking the local crowdedness into consideration when selecting mating parents. If a neighborhood is associated with too many solutions, the mating parents will be selected outside of this neighborhood. Likewise, \cite{LiZS19} proposed a self-adaptive mating restriction strategy that dynamically sets the mating selection probability according to the estimated quality of solutions associated with each subproblem. In particular, if a subproblem is frequently updated, it is assumed to be associated with a poor solution. Accordingly, it is assigned with a small mating selection probability and vice versa.

Since mating selection is mainly implemented within the neighborhood of each subproblem, it is anticipated that the size of each neighborhood can influence the balance of exploration versus exploitation (EvE). Instead of using a constant neighborhood size for each subproblem, \cite{ZhaoSZ12} and~\cite{ZhangZLZP17} proposed to maintain an ensemble of neighborhood size settings that compete with each other to implement an adaptive EvE balance, as depicted in~\pref{fig:selection}(b). Instead of using the neighborhood of each subproblem, an alternative is to use the neighborhood of solutions in the mating selection, such as clusters of solutions by using $k$-means algorithm~\cite{GholaminezhadI14,LiSZ19} or solutions close to each other in the objective space~\cite{WangZLZR18}.

\subsubsection{Environmental selection}
\label{sec:environmental_selection}

MOEA/D uses a steady-state environmental selection mechanism that the offspring solution is directly used to update the parent population right after its generation. The offspring can replace as many parent solutions as it can within the mating neighborhood. Such a greedy strategy is detrimental to the population diversity~\cite{LiZ09}. To mitigate this, it uses the same strategy as in the mating selection to give a small chance to update the subproblems outside of the mating neighborhood. To give a more robust fitness regarding a neighboring area, \cite{YuanXW14a} proposed an ensemble fitness ranking method to assign fitness value to each solution. It aggregates the ranks of scalarizing function values for a group of subproblems with regard to a solution.

Different from most environmental selection mechanisms that always prioritize the survival of non-dominated solutions against dominated ones, MOEA/DD~\cite{LiDZK15} is the pioneer that suggests emphasizing the survival of solutions lying in isolated regions with few neighbors even if they are dominated or infeasible ones, as depicted in~\pref{fig:selection}(c). The unpinned assumption is that these regions are likely to be under-exploited. Therefore, they are beneficial to the population diversity and help mitigate the risk of being trapped by local optima. Later, the importance of maintaining isolated solutions regardless of their convergence and feasibility is empirically validated in~\cite{ElarbiBS17}. Inspired by MOEA/DD, there have been many further attempts that try to allocate certain survival chances to solutions that are relatively inferior in terms of convergence but contribute more in terms of population diversity, e.g.,~\cite{XiangZLC17,AsafuddoulaSR18,HeZCZ19,ZhouCHX20}. In addition, MOEA/DD is an early work that aims to leverage the complementary effect of dominance- and decomposition-based selection strategies. Along this line, there have been some variants developed to exploit the benefits of combining various selection strategies under the same paradigm to achieve a better trade-off between convergence and diversity~\cite{LiuZYERS19}.

Instead of a steady-state scheme, \cite{LiZKLW14,WuLKZZ17} proposed the first generational environmental selection mechanism for MOEA/D. Its basic idea is to transform the environmental selection process as a matching problem among subproblems and solutions, where subproblems aim to find their appropriate solutions and vice versa, as depicted in~\pref{fig:selection}(d). In particular, the preference of subproblems is convergence while that of solutions is diversity. A stable matching among subproblems and solutions naturally achieves a balance between convergence and diversity. This stable matching-based selection mechanism inspires a series of follow-up MOEA/D variants that leverage the matching relationship between subproblems and solutions. For instance, \cite{LiKZD15} proposed another generational environmental selection mechanism that first builds the interrelationship among subproblems and solutions. Each solution is first attached to its most matching subproblem. Thereafter, the fittest solutions from those nonempty subproblems are chosen to survive to the next round while the remaining slot is filled with some less favorable solutions. This idea is extended as a global replacement mechanism that first finds the most matching subproblem for each offspring solution before updating the parent population~\cite{WangZGZ14,WangZZGJ16,WangZZGJ16a,ChenSZWS19}, a local niche competition~\cite{YuanXWZY16,YuanXWY16} and a correlative selection~\cite{LiuWBLJ21}.

\vspace{0.5em}
\begin{tcolorbox}[breakable, title after break=, height fixed for = none, colback = gray!40!white, boxrule = 0pt, sharpish corners, top = 4pt, bottom = 4pt, left = 4pt, right = 4pt, toptitle=2pt, bottomtitle=2pt]
    \faBomb\ Selection, i.e., survival of the fittest, is one of the most critical steps in both natural evolution and EC. In a nutshell, the major target of both mating and environmental selections is to achieve an EvE dilemma and also a balance between convergence and diversity. Key bottlenecks of the selection mechanisms in MOEA/D include: $\blacktriangleright$ Given the black-box nature of the underlying problem and also the dynamics of evolution, there is no rule of thumb to evaluate neither exploration nor exploitation (the same for the convergence and diversity). $\blacktriangleright$ Environmental selection is the springboard for analyzing the search dynamics of MOEA/D. However, related theoretical analysis has been significantly overlooked in the literature.

\end{tcolorbox}


\subsection{Reproduction Operators in MOEA/D}
\label{sec:reproduction}

Reproduction operators are used to generate offspring thus determining the way to explore the search space. MOEA/D~\cite{ZhangL07} applied the simulated binary crossover and polynomial mutation for offspring reproduction whereas the differential evolution considered in~\cite{LiZ09} has a much wider uptake thereafter. In principle, any reproduction operator proposed in the EC community can be used within MOEA/D. Here we review some representative operators being used in MOEA/D according to their types, as summarized in Table A$4$ of \textsc{Appendix}.

\subsubsection{Swarm intelligence (SI)}
\label{sec:swarm}

SI represents the collective intelligent behavior of decentralized, self-organized systems, such as groups of simple agents. It includes behaviors observed in nature, like social insects' foraging and cooperative tasks.
\begin{itemize}
	\item\underline{\textit{Particle swarm optimization (PSO)}}: Among the most popular SI methods, PSO has been widely adopted within MOEA/D~\cite{PengZ08,MoubayedPM10,MartinezC11}. They maintain a swarm of particles where each particle possesses its own position, velocity, and personal best vectors, alongside a population of global best vectors corresponding to different subproblems. The fitness of each particle is often assessed by a scalarizing function, e.g.,~\cite{MoubayedPM14,LinLDCM15,Pan0GW18,YuCGZYKZ18,XiangZCZ20}.
	
	\item\underline{\textit{Ant colony optimization (ACO)}}: MOEA/D-ACO marked the first ACO adaptation within MOEA/D~\cite{KeZB13}, assigning each ant to a subproblem. Ants in groups maintain a pheromone matrix to guide the search, constructing new solutions from both group pheromone and personal heuristic information. Further developments include using opposition-based learning~\cite{ZhangXLLS20}, negative pheromone~\cite{NingZSF20} and local optima avoidance~\cite{WuQJZX20}.	
\end{itemize}

Beyond PSO and ACO, other SI methods, such as artificial bee colony~\cite{ZhongXL14}, cuckoo search~\cite{ChenGLXCF19}, grey wolf optimization~\cite{MirjaliliSMC16} and whale optimization~\cite{Abdel-BassetMM21}, have been adapted to the MOEA/D framework.


\subsubsection{Model-based methods}
\label{sec:model}

Traditional reproduction operators in EC, including crossover and mutation as well as SI methods, rely on fixed heuristic rules or strategies. They primarily interact with individual solutions and seldom adapt to the evolving environment, which can rapidly change as evolution progresses. Additionally, the complex interdependencies between variables can twist or rotate the problem landscape. These potentially diminish the effectiveness of traditional operators due to their static nature and inability to learn from environmental changes. In response, a growing number of efforts have aimed to enhance reproduction operators with learning capabilities. The fundamental approach involves substituting heuristic operators with machine learning models to generate new candidate solutions. These models, trained on candidate solutions sampled from the current environmental context, offer a dynamic means to navigate the search space.
\begin{itemize}
	\item\underline{\textit{Estimation of distribution algorithm (EDA)}}: Its core idea involves estimating the distribution of promising candidate solutions by training machine learning models within the decision space. MEDA/D pioneered the use of multi-objective EDA based on decomposition for tackling the multi-objective traveling salesman problem (MOTSP)\cite{GaoZZ12,ZhouGZ13}. By decomposing a MOTSP into subproblems, MEDA/D assigns each subproblem with a probabilistic model incorporating both heuristic information and insights learned from the evolving population. New candidate solutions, or tours, are generated by sampling from these models. The effective approximation of the PF for MOTSP relies on the collaborative efforts of neighboring subproblems. This approach has been extended by using more expressive probabilistic models, such as probabilistic graphical models~\cite{desouza2015moeadgm}, population-based incremental learning~\cite{XingWLLQ17}, and cross-entropy~\cite{GiagkiozisPF14}.
	
	\item\ul{\textit{Covariance matrix adaptation evolution strategy (CMA-ES)}}: Initial attempts to incorporate CMA-ES into MOEA/D, such as those in~\cite{ParkL13} and~\cite{MartinezDLBAT15}, involved associating each subproblem with a unique Gaussian model, leading to exponentially increased computational complexity with population size. MOEA/D-CMA~\cite{LiZD17} addresses this by adopting a hybrid strategy that combines differential evolution with CMA-ES, optimizing one subproblem at a time within a group.
	
	\item\underline{\textit{Probability collectives (PC)}}: Originating from game theory and statistical physics~\cite{WolpertSR06}, PC emphasizes manipulating probability distributions rather than constructing point-wise models. MOPC/D~\cite{MorganWC13} uses this approach, along with local search, to build distributions centered around the PS's neighborhood.
		
	\item\underline{\textit{Pareto set model}}: Focused on capturing regularities through learning techniques, this strategy guides evolutionary search towards promising offspring solutions. ~\cite{ZhouZZZ19} initially approximated the PS using a linear model, while~\cite{WangZZF23} later employed a generative topographic mapping to navigate offspring reproduction by understanding the population's manifold distribution.
\end{itemize}

\subsubsection{Adaptive operator selection and hyper-heuristics}
\label{sec:aos}

Different reproduction operators have their unique characteristics and search dynamics, underscoring the challenge that no single operator can universally excel across a diverse array of problems. Addressing the EvE dilemma requires a dynamic adjustment strategy that favors exploration in the early stages of problem-solving and shifts to exploitation as the search converges on the optimum. Adaptive operator selection (AOS) emerges as a solution, dynamically selecting the optimal reproduction operator based on real-time feedback. MOEA/D-FRRMAB~\cite{LiFKZ14} is the first instance that applies AOS for multi-objective optimization. Based on FRRMAB, many other AOS variants have been developed afterward (e.g.,~\cite{GoncalvesAP15,QiBMML16,GoncalvesPAKVD17,SunL20}). In addition to AOS, \cite{LiFK11,LinLYDCLWC16,CuiLLCL16,LinTMDLCM17} also consider adaptively controlling the parameters associated with the reproduction operator in order to achieve the best algorithm setup. Rather than adaptively selecting the \lq appropriate\rq\ reproduction operator on the fly, another line of research (e.g.,~\cite{WangYLZWCC19,XieQWY20,LuoLJJC21}) is to build an ensemble of reproduction operators and use them simultaneously.

Different from AOS, hyper-heuristics (HH) offer a high-level strategy for heuristic selection and generation, capable of creating new operators that evolve with the optimization process. Early HH works within the MOEA/D framework, such as~\cite{GoncalvesKAV15a,GoncalvesKAV15b,PrestesDGAP17,FerreiraGP17}, mainly use a variant of FRRMAB for heuristic generation or selection. HH's integration with additional heuristic strategies, such as irace~\cite{PrestesDLGA18} and bandit models~\cite{GoncalvesALD18,AlmeidaGVLD20}, a two-level approach for complex problems.

\subsubsection{Mathematical programming methods}
\label{sec:mathematical}

Reproduction operators, as previously discussed, are designed to adaptively manage the EvE dilemma throughout the evolutionary process. However, they often excel more in global search scenarios. To enhance exploitation in specific promising regions, hybrid methods that integrate mathematical programming techniques---both gradient-based and derivative-free---into MOEA/D have been explored. Among these, the Nelder and Mead's method~\cite{NelderM65}, a popular derivative-free optimization approach that conducts a nonlinear simplex search through reflection, expansion, and contraction, has been widely applied within MOEA/D~\cite{MartinezMC11,MartinezC12,MartinezC13,MartinezC16,PrestesAG15,ZhangZZS17}. Additionally, the Karush-Kuhn-Tucker optimality theory-based proximity measure, KKTPM~\cite{DebA16}, originally developed to assess the convergence of non-dominated solutions to the PS, has been utilized as a key component or termination criterion in local search strategies within NSGA-III~\cite{DebAS17,AbouhawwashSD17,SeadaAD19}.

\subsubsection{Memetic search methods}
\label{sec:memetic}

Memetic algorithm~\cite{MoscatoN92}, an extension of the traditional genetic algorithm, incorporates individual learning or local search techniques to reduce the risk of premature convergence. Within MOEA/D, memetic algorithms have been tailored for combinatorial MOPs, offering a distinct approach from the mathematical programming-based local search. We briefly discuss several memetic MOEA/D variants based on their local search strategies.
\begin{itemize}
    \item\underline{\textit{Pareto local search (PLS)}}: It focuses on exploring solution neighborhoods to identify new non-dominated solutions or better local optima~\cite{LustT10}. MOMAD was the pioneering integration of PLS with MOEA/D, utilizing an auxiliary population for the PLS process~\cite{PaqueteS03}. Shi et al.~\cite{ShiZS20} enhanced PLS efficiency through parallel computation and problem decomposition, significantly accelerating the PF approximation process.

	\item\underline{\textit{Neighborhood search (NS)}}: NS searches optimal or near-optimal solutions by exploring a current solution's neighborhood, using diverse operators for varied neighborhood sizes or behaviors~\cite{HuWLWLY19,CaiHGGZFH19,ZhouKCWLHW20,WangYZGSZ21}. NS has been integrated into MOEA/D to enhance solution quality through different neighborhood dynamics~\cite{AlhindiZ14,ZhouWWW18,DuXZCH19,AlhindiAAATA19}.

    \item\underline{\textit{Simulated annealing (SA)}}: SA is a technique inspired by statistical physics~\cite{KirkpatrickGV83}, effective for combinatorial optimization and incorporated into MOEA/D for improved solution exploration and optimization~\cite{LiS08,LiL11}. Later, the same idea was applied for solving software next release~\cite{CaiCFGW17} and hybrid flow shop problems~\cite{JiangZ19a}.

    \item\underline{\textit{Greedy randomized adaptive search procedure (GRASP)}}: It is merged with MOEA/D to form a hybrid evolutionary meta-heuristic that leverages GRASP and path-relinking for enhanced search efficiency~\cite{KafafyBB11,AlhindiZT14}.

    \item\underline{\textit{Other local serach}}: In addition to the classic local search, there are some other local search operators designed by considering bespoke problem knowledge. For instance,\cite{LeiSY18} proposed two local search operators for exam tabling problems. One is called period-supplement local search that improves the distribution of non-dominated solutions. The other is called bi-directed local search operator that improves the quality of the solutions. \cite{GharaeiJ18} proposed a path solution representation along with several neighborhoods and local search operators for multi-factory supply chain problems.
\end{itemize}

\begin{tcolorbox}[breakable, title after break=, height fixed for = none, colback = gray!40!white, boxrule = 0pt, sharpish corners, top = 4pt, bottom = 4pt, left = 4pt, right = 4pt, toptitle=2pt, bottomtitle=2pt]
    \faBomb\ Key bottlenecks of the existing subproblem formulations include: $\blacktriangleright$ Although many reproduction operators have been applied in MOEA/D, there does not exist one winner takes all in view of the no-free-lunch theorem. In other words, a reproduction operator might only excel in one type of problems or the heuristics even need to be tailored according to the problem characteristics. $\blacktriangleright$ Although both AOS and HH are expected to autonomously identify the bespoke operator, they are essentially an online optimization problem, which is challenging to solve on top of the underlying MOP.
\end{tcolorbox}


\section{Advanced Topics}
\label{sec:advanced}

This section plans to review selected important developments of MOEA/D variants for four specific scenarios.

\subsection{Constraint Handling}
\label{sec:constraint}

Real-life optimization scenarios inherently come with various constraints. However, constraint handling has surprisingly been ignored not only in the seminal paper of MOEA/D~\cite{ZhangL07}, but also been lukewarm until~\cite{LiCFY19}. Constraints not only create infeasible regions in the search space, posing obstacles to effective evolutionary search but can also distort the PF in terms of its shape and integrity. The current constraint handling approaches fall into three categories, as summarized in Table A$5$ of \textsc{Appendix}. Note that this categorization is not limited within MOEA/D, but is general to any EMO.

\subsubsection{Feasibility-driven methods}
\label{sec:feasibility}

As the first endeavor in this domain, \cite{JanZ10} incorporated penalty terms within the scalarizing function to favor feasible solutions. Later, g-DBEA~\cite{AsafuddoulaRSA12,AsafuddoulaRS15} developed an adaptive method to calculate constraint violations, tailoring the violation threshold to the constraints, size of the feasible region, and search outcomes. The successful constraint-domination principle from NSGA-II~\cite{DebAPM02} was also adapted for NSGA-III variants~\cite{JanK13,JainD14}. In~\cite{LiDZK15}, a novel approach was proposed considering the prioritization of an infeasible solution over feasible ones if located in an isolated subregion. Inspired by early results in~\cite{YangCF14}, the $\epsilon$-constraint method was modified for MOEA/D, with variants such as~\cite{FanLCHFYMWG19} dynamically adjusting the $\epsilon$ level based on the proportion of feasible solutions in the population. In~\cite{FanLCLWZDG19,FanWLYYYSR20}, a push and pull search framework was proposed. Initially, the push stage explores the feasible region; followed by the pull stage, where $\epsilon$-constraint is used to navigate the search within the infeasible region.

\subsubsection{Weight vector adaptation}
\label{sec:adatpation_constraint}

The second category focuses on adapting the distribution of weight vectors to the PF distortion caused by constraints, as those introduced in~\pref{sec:weight}. The core idea involves progressively identifying and removing weight vectors targeting infeasible parts of the objective space, while introducing new ones in feasible areas, e.g.,~\cite{ChengJOS16} and~\cite{AsafuddoulaSR18}. \cite{PengLG17} proposed maintaining two types of weight vectors to explore both feasible and infeasible regions. It uses infeasible weights to preserve a set of well-distributed infeasible solutions, thereby utilizing valuable information from these regions. These weights are dynamically adjusted throughout the evolution process to favor individuals with better objective values and smaller constraint violations.

\subsubsection{Trade-off among convergence, diversity and feasibility}
\label{sec:tradeoff}

Traditional constraint handling methods prioritize driving the population toward feasible regions before addressing the balance between convergence and diversity. This approach risks in local optima or feasible regions, especially when these regions are narrow or widely separated in the search space. C-TAEA~\cite{LiCFY19} is the pioneering work incorporating this three-way trade-off into its algorithm design. It maintains two complementary archives: a convergence archive to guide the population towards the PF, and a diversity archive to explore underutilized areas, including infeasible regions. This concept has sparked many follow-up studies using multiple populations for constraint handling, (e.g.,~\cite{TianZXZJ21,JiaoZLYO21,LiangQYQYGW23,XieLWWPJ23,ZouSLHYZL24}). More recently, Li et al., proposed tackling constrained multi-objective optimization problems without accessing the constraint violation~\cite{LiLL22,LiLLY24,WangL24}.

\subsection{Optimization in Dynamic and Uncertain Environments}
\label{sec:dynamic}

In addition to the challenge of highly constrained search spaces, another complexity of real-world black-box optimization problems comes from dynamic and uncertain environments. These conditions lead to time-varying objective functions, resulting in temporally evolving PSs and PFs. In principle, prevalent approaches in the realm of dynamic multi-objective optimization can be readily adapted to MOEA/D. Among them, the most popular one is the use of predictive models to estimate the seeding population after environmental changes. For instance, \cite{MurugananthamTV16} proposed leveraging the Kalman filter to predict the evolving trajectories of PSs and PFs with time. This concept has been further developed to address unknown dynamic environments through inverse modeling~\cite{GeeTA17a} and autoregressive models~\cite{CaoXGL19,CaoXGBZ20,RambabuVTJ20}. Besides, MOEA/D has the unique capability to tailor its search directions to evolving environments by managing the distribution of weight vectors. A representative work along this line is~\cite{ChenLY18} that pioneered the idea of dynamic multi-objective optimization with a changing number of objectives. This idea has been explored in~\cite{HuZZYOW20,AhrariESEC21,PengMZLZW23}. Additionally, transfer optimization has been another popular method that selectively leverages the historical data to boost the evolutionary search in the new environment either by seeding a new initialized population~\cite{JiangWQGGT21} or informing the adjustment of weight vectors~\cite{JiangWHY21,LinYMJT24}. Last but not least, given the complexities of uncertain environments, employing an ensemble of dynamic response strategies offers resilience and versatility~\cite{LiangZZY19,LiuLJJ21,JiangGCL22}.

\subsection{Surrogate-Assisted Methods for Expensive Optimization}
\label{sec:expensive}

Real-world problems are characterized by time- or resource-intensive physical experiments or numerical simulations. For instance, evaluating a single function based on computational fluid dynamics simulations can take anywhere from minutes to hours. Traditional evolutionary algorithms, which require a large number of function evaluations, become impractical. To address this, surrogate models have emerged as a widely adopted solution, serving as efficient proxies for expensive objective functions, as summarized in Table A$6$ of \texttt{Appendix}.

ParEGO~\cite{Knowles06} is the first one that introduces surrogate modeling into the decomposition-based EMO algorithms, even earlier than the canonical MOEA/D~\cite{ZhangL07}. It aggregates a MOP into a scalar function using an augmented Tchebycheff function and a randomly sampled normalized weight vector for each iteration. This method extends the efficient global optimization (EGO) strategy~\cite{JonesSW98}, where a Gaussian process (GP) model represents the expensive objective function, and the search is driven by maximizing expected improvement based on the GP model, to address expensive MOPs. Four years later, Zhang et al.~\cite{ZhangLTV10} proposed MOEA/D-EGO, adapting EGO for MOEA/D. Given the co-evolution of a population of subproblems in MOEA/D, building a separate surrogate model for each would be computationally prohibitive. Instead, MOEA/D-EGO builds a GP model for each objective, adapting EGO for subproblem optimization using specific mean and variance formats related to weighted sum and Tchebycheff functions. K-RVEA~\cite{ChughJMHS18}, addresses computationally expensive many-objective optimization by utilizing the uncertainty estimates from Kriging models. It adapts the distribution of weight vectors and the positioning of solutions in the objective space to balance convergence and diversity. 

There have been various variants by introducing other machine learning models as surrogates, such as radial basis function~\cite{MartinezCZ13,MartinezC13}, extreme learning machines~\cite{PavelskiDAGV14,PavelskiDAGV16}, multi-level surrogate models~\cite{ZhangLHJ17}, classifiers~\cite{HeCJ019}, support vector machines~\cite{SonodaN20}, and Walsh functions~\cite{PruvostDLVZ20}. In addition to maximizing expected improvement, new infill criteria have been developed, such as bi-objective formulations~\cite{FengZZTYM15}, dynamic weighted aggregations~\cite{WangJSO20}, and closed-form improvements for PBI/IPBI functions~\cite{NamuraSO17}. Efforts also include leveraging latent information of surrogate models~\cite{ChenL23,LiC23}, complementary infill criteria~\cite{LiGGSH21} and the use of multiple surrogate models~\cite{HabibSCRM19}. To tackle the curse of dimensionality, dimensionality reduction like principal component analysis~\cite{ZhaoZCZYSHY20} and feature selection~\cite{TanW20} have been employed. Moreover, the advent of transfer learning has facilitated knowledge sharing and reuse within MOEA/D, such as~\cite{MinOGG19} and~\cite{YangDJC20}. Leveraging MOEA/D's piecewise linear assumption and the similarity among neighboring subproblems, \cite{LuoGOW19} proposed multi-task GP models for collaborative surrogate modeling of representative subproblems. Recently, Chen and Li proposed a transfer evolutionary optimization framework for designing data-driven evolutionary optimization methods based on transfer learning~\cite{ChenL21,LiCY24}.

\subsection{Preference Incorporation in MOEA/D}
\label{sec:preference}

The methodologies reviewed thus far aim to approximate the entire PF. However, as~\cite{LiDY18,LiLDMY20,LiNGY22} discussed, presenting decision-makers (DMs) with a wide array of trade-off alternatives not only burdens the DM but can also introduce irrelevant or noisy information into the decision-making process. Over the past two decades, there has been an increasing trend towards integrating EMO and MCDM to incorporate DM preferences directly into the evolutionary search~\cite{WangOJ17,XinCCIHL18,LiLDMY20,YangL23}. This approach focuses on approximating the region of interest (ROI), which is often just a partial region of the PF. In MOEA/D, the subproblems' configuration, determined by the distribution of weight vectors, influences the location of approximated Pareto-optimal solutions. Therefore, integrating DM preferences into MOEA/D involves adjusting weight vector distribution based on the DM's provided preference information. As outlined in Table A$7$ of \texttt{Appendix}, we will review current research in MOEA/D's preference incorporation, categorized by the preference elicitation method, whether \textit{a priori} or \textit{interactively}.

\subsubsection{A priori preference elicitation}
\label{sec:a_priori}

The basic strategy for incorporating \textit{a priori} preference information into MOEA/D involves biasing the distribution of weight vectors based on the DM's preferences. Weight vectors, initially distributed evenly, are adjusted to focus the search towards the ROI. For instance, \cite{LiCMY18} introduced a method named NUMS that maps evenly distributed weight vectors to positions closer to the DM's reference point on a simplex. Unlike this static approach, other studies~\cite{MohammadiOL12,MohammadiOLD14,ZhuHJ16,XuLA021,XuLA21,XuLA22,XuLL24,XuLL24a} dynamically shift weight vectors towards the ROI as the population evolves, favoring solutions near the DM's reference point. Further, approaches like a-PICEA-g~\cite{WangPF13} and cwMOEA/D~\cite{PilatN15} co-evolve the population and weight vectors, approximating the ROI simultaneously. a-PICEA-g generates a new population of weight vectors randomly in each iteration, while cwMOEA/D employs a mutation operator to create new candidate weight vectors. Another method involves defining a preference region, usually around a DM-provided reference point, within a specified boundary (e.g., a tolerance angle~\cite{MeneghiniG17} or a radius~\cite{MaLQLJDWDHZW16}), from which new weight vectors are sampled. In~\cite{YuZSL16}, a scaled simplex was used to specify the preference region, starting with biased weight vectors at the simplex's corners and iteratively adding weight vector pairs inside of the simplex.

\subsubsection{Interactive preference elicitation}
\label{sec:interactive}

Differently, \textit{interactive} preference elicitation allows the DM to progressively learn about the problem characteristics and adjust their preferences accordingly. One of the earliest works in this area, PMA~\cite{Jaszkiewicz04}, periodically engages the DM to rank pairs of solutions, constructing a set of linear value functions for environmental selection and ROI approximation. iMOEA/D~\cite{GongLZJZ11} proposed a virtual interaction mechanism, where the DM periodically selects promising solutions to refine the utility function and adjust weight vectors. \cite{LiCSY19} developed an interactive framework within MOEA/D. It consists of three modules, i.e., consultation, preference elicitation, and optimization. Specifically, after every several generations, the DM is asked to score a few candidate solutions in a consultation session. Thereafter, an approximated value function, which models the DM's preference information, is progressively learned from the DM's behavior. In the preference elicitation session, the preference information learned in the consultation module is translated into the form that can be used in a decomposition-based EMO algorithm, i.e., a set of reference points that are biased toward the ROI. Most recently, this framework was generalized to Pareto- and indicator-based EMO, and the preference is learned from implicit information~\cite{LiLY23}. In~\cite{TomczykK20}, IEMO/D proposed employing Monte Carlo simulations and rejection sampling to model the DM's preferences.


\section{Emerging Topics and Future Directions}
\label{sec:others}

In this section, we explore some emerging topics that transcend the realm of decomposition-based EMO, reflecting future research directions for our EMO community.

\subsubsection{Large-scale MOPs (LSMOPs)} 

Real-world optimization problems often involve a large number of decision variables, known as LSMOPs. One of the most recognized approaches in the EC community for tackling LSMOPs is cooperative coevolution. Its basic idea is to decompose the original problem into smaller, manageable subproblems based on decision variable analysis (DVA) that identifies variable interactions. This allows for a collaborative solution process. As the first attempt, \cite{MaLQWLJYG16} introduced a DVA technique that categorizes decision variables by their control properties—either convergence- or spread-related. This approach analyzes the interaction structures of these variables separately, employing random permutations. Further developments include extending this DVA technique into a distributed solution approach~\cite{CaoZLL17}, a graph representation~\cite{CaoZGLM20}, and an adaptive localized version~\cite{MaHYWW22}. Another strategy~\cite{QiBMML16} involves a self-adaptive operator selection mechanism, allowing MOEA/D to address LSMOPs without variable interaction analysis. However, compared to real-world scenarios like the GPT-$4$, which contains $1.76$ trillion parameters, current EMO algorithms for LSMOPs---handling usually less than $10,000$ decision variables---fall short of industrial-level requirements. To elevate the scalability of EMO, we need to investigate novel problem transformation and dimension reduction techniques, which can compress the high-dimensional search space, making optimal solutions achievable with an acceptable number of function evaluations.

\subsubsection{Multimodal MOPs (MMOPs)}

In some real-world optimization scenarios, such as space mission design problems~\cite{SchutzeVC11}, diesel engine design problems~\cite{HiroyasuNM05} and rocket engine design problems~\cite{KudoYF11}, there exist some dissimilar solutions in the decision space but lead to similar or equivalent quality in the objective space, as known as MMOPs. Solving an MMOP aims to locate (almost) equivalent Pareto-optimal solutions as many as possible. To maintain the diversity of the population in both the objective and decision spaces, \cite{TanabeI18} and~\cite{TanabeI20a} proposed a general framework that uses the environmental selection mechanism of the baseline EMO algorithm to maintain diversity in the objective space while the diversity in the decision space is maintained by a simple niching criterion. Recently, a graph Laplacian based optimization assisted by decomposition in both decision and objective spaces is proposed in~\cite{PalB21} for solving MMOPs. Very recently, \cite{WeiGGY24} proposed transforming a MMOP into a bi-objective optimization problem, thus promoting convergence in the objective space while maintaining diversity in the decision space. Future developments are expected to focus on more effective distance metrics and diversity preservation strategies to better identify multiple equivalent Pareto-optimal solutions simultaneously.

\subsubsection{Variable-length MOPs}

In conventional optimality theory, the length of decision variables is always assumed to be fixed. However, it is not uncommon that the optimal length of a decision variable is unknown \textit{a priori}, let alone the optimum, in many real-world applications, such as composite laminate stacking problem~\cite{SoremekunGHW01}, sensor coverage problem~\cite{TingLCW09} and wind farm layout problem~\cite{HerbertPRLC14}. This can lead to a variable-length Pareto structure where the number of variables in two Pareto-optimal solutions might be different. One of its key challenges is the varying length of variables that disrupt the genetic reproduction operators in EA. Furthermore, due to the existence of variable-length Pareto structure, approximating the PF can be biased as the local region with a larger variable size is more difficult than those with smaller variable sizes~\cite{LiD17}. To address these issues, MOEA/D-VLP~\cite{LiDZ19} proposed a two-level decomposition strategy where the global level is the same as the conventional MOEA/D while the local level decomposes an MOP into multiple MOPs with different ranges of dimensionality to approximate various local PF segments. In addition, an adaptive strategy is proposed to allocate computational efforts towards different parts of the PF to mitigate the bias towards any relatively easier PF regions. Recently, \cite{MaLYWMWJT23} proposed using a simple dimensionality incremental learning strategy to choose representative solutions. As a next step, it is desirable to develop bespoke subproblem formulations for different regions of the PF and a balanced way to strategically allocate computational budget.

\subsubsection{Sparse optimization}

Sparse optimization problems widely appear in many real-world applications such as data mining, variable selection, visual coding, signal and image processing. Its ultimate goal is to find a sparse representation of a system, which usually aims to minimize a data-fitting term and a sparsity term simultaneously. As a dominating strategy in sparse optimization, the classic iterative thresholding often aggregates these two terms into a single function, where a relaxed parameter is used to balance the error and the sparsity. In particular, the choice of this relaxed parameter is sensitive to the performance of the iterative thresholding methods. There have been some initial attempts~\cite{LiSXZ12,LiSWZ18,LiZDX18} that transform the sparse optimization problem as a MOP and they apply tailored MOEA/D to solve it. Due to the unknown sparsity level of the underlying problem, it is challenging to maintain the diversity of the population. In addition, it is imperative to design bespoke performance indicators to measure the convergence, diversity, and sparsity of the population simultaneously.

\subsubsection{Parallel computing}

In principle, the decentralized nature of decomposition is ideal for distributed computing resources. However, integrating parallelism into decomposition approaches presents challenges, given the heterogeneous and complex nature of modern large-scale computing platforms. It requires that parallelism be tightly integrated with decomposition components, rather than treated as a separate entity. There have been limited studies on distributed MOEA/D implementations, such as the classic master-slave model~\cite{LiaoIPS20} and GPU-based CUDA technology for real-time MOPs~\cite{YuLQ18}. As for a future direction along this line, developing a mapping function to align decomposed subproblems with the parallelism offered by the computing platforms is crucial. Additionally, designing adaptive distributed algorithms that cater to the specific characteristics of subproblems is necessary.

\subsubsection{Multi-agent system vs MOEA/D}

In theory, the decomposed subproblems position MOEA/D akin to multi-agent systems, where each subproblem functions as an autonomous agent engaging in a cooperative search, within a pre-defined neighborhood, for its optimal solution through population-based meta-heuristics. This analogy has been explored by the matching-based selection mechanism~\cite{LiZKLW14,LiKZD15,WuLKZZ17}, while the full potential of this concept has unfortunately yet been fully exploited in literature. Another light touch in~\cite{MorganWC13} applied probability collectives~\cite{WolpertSR06}---originally proposed for distributed optimization---to MOEA/D in a multi-agent system manner. Given the wealth of theoretical and methodological advancements in multi-agent systems and game theory, there lies a promising avenue to harness these insights for crafting more effective and principled EMO algorithms.

\subsubsection{Theoretical studies}

In our EMO community, rigorous theoretical analysis has often lagged behind methodological developments. There are two primary types of theoretical studies within the MOEA/D framework. One mainly focused on the relationships among classic subproblem formulations and the impacts of their respective hyperparameters. For instance, \cite{MaZTYZ18} theoretically explored the geometric properties of TCH, uncovering asymptotic equivalences with the Hypervolume indicator. Similarly, \cite{SinghD20} investigated the equivalence between PBI and AASF under various hyperparameter settings. Another line of research concerns the runtime analysis of MOEA/D~\cite{HuangZCHLX21,HuangZLL21} and NSGA-III~\cite{WiethegerD23}. However, these studies are often limited to \lq overly simplified\rq\ synthetic benchmark problems with known Pareto-optimal solutions. 


\section{Concluding Remarks}
\label{sec:conclusions}

MOEA/D bridges the gap between population-based meta-heuristics, EMO in particular, and conventional MCDM. This article has provided a comprehensive survey of the developments of MOEA/D to date. We began with a gentle tutorial on MOEA/D's basic workings to ensure the survey is self-contained. Our survey then elaborated on the four core design components: weight vector settings, subproblem formulations, selection mechanisms, and reproduction operators. We also covered selected advanced topics, such as constraint handling, optimization in computationally expensive scenarios, and preference incorporation. Finally, we highlighted emerging directions for future development. Note that these areas are beyond the scope of MOEA/D, but are of universal significance for our entire EMO community.

\begin{tcolorbox}[breakable, title after break=, height fixed for = none, colback = gray!40!white, boxrule = 0pt, sharpish corners, top = 4pt, bottom = 4pt, left = 4pt, right = 4pt, toptitle=2pt, bottomtitle=2pt]
    \faCoins\ \underline{\textsc{Resources}}: There are a series of technical activities themed on the decomposition multi-objective optimization. For instance, there is a dedicated \href{https://sites.google.com/view/moead/}{website} gathering various resources of MOEA/D, including papers, source codes and active researchers. Sponsored by the IEEE Computational Intelligence Society, there has been a \href{https://cola-laboratory.github.io/docs/misc/dtec/}{Task Force} themed on decomposition-based techniques in EC since 2017. This is an international consortium that brings together global researchers to promote an active state of this area. Since 2018, there have been tutorials, workshops and special sessions regularly organized within major EC conferences such as PPSN, GECCO, CEC and SSCI.

\end{tcolorbox}

\section*{Acknowledgment}
This work was supported in part by the UKRI Future Leaders Fellowship under Grant MR/S017062/1 and MR/X011135/1; in part by NSFC under Grant 62376056 and 62076056; in part by the Royal Society under Grant IES/R2/212077; in part by the EPSRC under Grant 2404317; in part by the Kan Tong Po Fellowship (KTP\textbackslash R1\textbackslash 231017); and in part by the Amazon Research Award and Alan Turing Fellowship.

\bibliographystyle{IEEEtran}
\bibliography{moead}

\end{document}